\documentclass{article}

\usepackage{arxiv}

\usepackage{hyperref}       % hyperlinks
\usepackage{url}            % simple URL typesetting
\usepackage{booktabs}       % professional-quality tables
\usepackage{amsfonts}       % blackboard math symbols
\usepackage{nicefrac}       % compact symbols for 1/2, etc.
\usepackage{microtype}      % microtypography
\usepackage{lipsum}		% Can be removed after putting your text content
\usepackage{graphicx}
\usepackage{natbib}
\usepackage{doi}
\usepackage[T2A,T1]{fontenc}
\usepackage[utf8]{inputenc}
\usepackage[russian, english]{babel}

\title{Fine-tuning GPT-3 for Russian Text Summarization}

\author{ \href{https://orcid.org/0000-0002-5643-9245}{\includegraphics[scale=0.06]{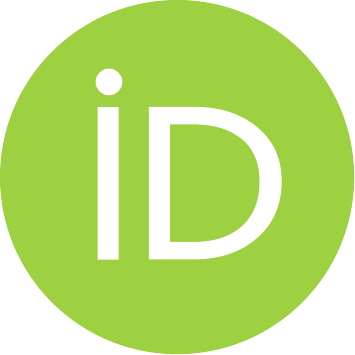}\hspace{1mm}Nikolich Alexandr} \\
	Federal State Budget Educational Institution of Higher Education «MIREA - Russian Technological University»\\
	Moscow, Russia \\
	\texttt{nikolich.a.d@edu.mirea.ru} \\
	\And
	\href{https://orcid.org/0000-0001-7482-3040}{\includegraphics[scale=0.06]{orcid.pdf}\hspace{1mm}Puchkova Arina} \\
	National Research University Higher School of Economics\\
	Moscow, Russia\\
	\texttt{adpuchkova@edu.hse.ru} \\
}

\renewcommand{\shorttitle}{Fine-tuning GPT-3 for Russian Text Summarization}

\hypersetup{
pdftitle={A template for the arxiv style},
pdfsubject={q-bio.NC, q-bio.QM},
pdfauthor={David S.~Hippocampus, Elias D.~Striatum},
pdfkeywords={First keyword, Second keyword, More},
}

\begin{document}
\maketitle
\selectlanguage{english}

\begin{abstract}
	Automatic summarization techniques aim to shorten and generalize information given in the text while preserving its core message and the most relevant ideas. This task can be approached and treated with a variety of methods, however, not many attempts have been to produce solutions specifically for the Russian language despite existing localizations of the state-of-the-art models. In this paper, we aim to showcase ruGPT3 ability to summarize texts, fine-tuning it on the corpora of Russian news with their corresponding human-generated summaries. Additionally, we employ hyperparameter tuning so that the model’s output becomes less random and more tied to the original text. We evaluate the resulting texts with a set of metrics, showing that our solution can surpass the state-of-the-art model’s performance without additional changes in architecture or loss function. Despite being able to produce sensible summaries, our model still suffers from a number of flaws, namely, it is prone to altering Named Entities present in the original text (such as surnames, places, dates, etc.), deviating from facts stated in the given document, and repeating the information in the summary.
\end{abstract}

% keywords can be removed
\keywords{text summarization \and rugpt3 \and russian language}

\section{Introduction}

Automatic summarization seeks to present given information in a more compact form, determining the key messages of the text and eliminating unnecessary details and filler sentences. Such a task can be applied to various domains and informational sources. For illustration, numerous studies have been dedicated to summarizing news, lectures, and medical data, with some authors utilizing not only lexical but also acoustic and visual data [1].  \\

	In general, text summarization methods can be broadly divided into extractive and abstractive approaches. The extractive one aims to select the most relevant sentences or tokens from the original document and comprise them into a summary. The abstractive approach, on the other hand, generates a summary ‘from scratch’ – it is conditioned on the source text but is capable of producing novel words, rearranging sentences, paraphrasing, replacing tokens with corresponding synonyms, and so forth. The advantages and disadvantages of this are obvious: generative models can ensure a higher level of linguistic fluency and provide more a coherent narrative imitating human speech, however, they can significantly alter the content of the original texts and distort the factual details. \\
	
Nonetheless, in current settings with the advances of sequence-to-sequence frameworks, abstractive models tend to outperform extractive ones. This holds true even in the cases of very limited supervision, with as little as 1000 training documents and respective summaries being provided to the model [2]. Most of this success can be attributed to the breakthrough of Transformer-based architectures [3]: due to their ability to ‘understand’ the logic of natural language, they can better decipher the relationship between sentences and the constituent words, thus, determining the general idea of the input texts. \\

Despite Transformers being widely used in recent works on text summarization in English, not many efforts have been made in order to apply them to the Russian language. To the best of our knowledge, current solutions employ Universal Transformer architecture [4] and BART-like models [5] to perform this task. Even so, BART is not exactly a generative model, and it is trained to reconstruct a document from its noisy version rather than put together text sequences on the fly. ruGTP3, however, is already capable of producing summaries and paraphrasing just as the original GPT3 model. We propose that with the additional fine-tuning on the contents of news articles and their corresponding summaries provided in Gazeta dataset [6] ruGPT3 can learn to deliver adequate summaries. Additionally, we experimented with hyperparameter-tuning in order to reduce the stochasticity of the text generation and penalize the repetition of the information. We calculate several metrics of text summarization quality alongside the human evaluation and conclude that our approach yields satisfactory results that can outperform the previous solutions. Our model, however, tends to produce a number of repeated errors that are yet to be fixed. 

\section{Related work}

First attempts of automatic text summarization utilized Recurrent Neural Networks [2]. However, the introduction of the self-attention mechanism Transformer architecture allowed to significantly improve the quality of many NLP tasks’ solutions ensuring that the model learns the logic of the natural language and semantics of the words. Rush et al. [7] were the first to use these mechanisms for abstractive text summarization. Recent works in this field that employ Transformer-based architectures rely on several techniques. First, one can train a masked model or denoising autoencoders that can generate text as if restoring or filling the gaps in the original. Current solutions feature T5 network that was trained on corrupted texts (with deletions, replacement, or an excess of characters) with various spans of masking [8]. Another popular model used for summarization is BART [9] that was trained to restore the original texts after arbitrary noising functions were applied to them. Perhaps the most advanced model of text summarization, PEGASUS [2],  follows the same logic: key sentences are being fully masked while the model is trained to restore them. \\

	Another approach is to use purely generative models such as GPTs knowing that training on the large corpora makes them familiar with the task of summarization. For example, Radford et al. [10] show that a large Transformer trained on texts from the Internet is capable of generating summaries after the prompt ”TL;DR” (‘too long; didn’t read’ – this phrase usually precedes the summary of a post on the Internet forums). Analogously, ruGPTs can produce summaries after the prompt \selectlanguage{russian}{‘короче’}
	(‘briefly’). \\
	
\selectlanguage{english} Several works employ GPTs for summarization of English texts [11], but to the best of our knowledge, no such experiments were done in regards to Russian. There is one work on fine-tuning ruGPT3 for a similar task of sentence simplification [12] that yields high values of the automatic quality metrics. The current state-of-the-art Russian summarization model, however, builds upon fine-tuning multilingual BART [6]. 

\section{Dataset}

Our model was trained on  Gazeta dataset [6]. It contains 63435 articles published by gazeta.ru newsagency from June of 2010 up to March of 2020 with their respective URLs, dates of posting, titles, and summaries. After scraping the source, the author also cleaned the data by removing the summaries that are too short or too long and those that have too narrow or too wide n-gram overlap with the text. \\

	The key advantage of this dataset is that it is one of the few datasets for summarization in Russian that provide actual summaries and not just headlines. While there is an abundance of data suitable for headline generation it is a separate task from text summarization since the summarization models aim not only to detect the main thought of the text but also to provide some necessary details. 
	For training, we split the dataset roughly in 90:10 proportion with 57665 examples belonging to a training set and another 5770 used for testing. 

\section{Model Description}

For our experiments, we used a pretrained ruGPT3Small model (125M parameters, 2048 context length, 1024 sequence length) provided by Sber [13]. It was trained on 80B tokens for 3 epochs with total training time being around one week when using 32 GPUs. \\

	The text-summary pairs were fed into the model with addition of special tokens in the following way: ‘<s>Text:{text} <|sep|> Summary:{summary} </s>’  so that <|sep|> separates the source text from the target summary. During the model inference the input is also passed as 'Text:{text} <|sep|> Summary:' so that the model generates the summary as a continuation of the prompt.
	
\section{Experiments and hyperparameter tuning}

It is known that generative models such as ruGPT3 can deviate significantly from the source text, altering the details or even going onto a completely different subject. However, the randomness of the model’s prediction can be somewhat controlled by hyperparameters such as temperature. The temperature of the GPT model ranges from 0 to 1, and generally with a lower temperature the model is more likely to choose tokens with a higher probability of occurrence (vice versa, the higher the temperature, the more ‘creative’ the model gets with its generation). For our task, we set the temperature to 0 so the model becomes more restricted to the contents of the source text. \\

	Apart from that, model performance was further improved by tuning the beam search algorithm. The top-k method picks the most probable words, then selects top-k most probable words that follow the initial k words, and repeats the process generating numerous sequences (or beams), selecting the one with the highest compound likelihood as an output in the end. The top-p technique restricts the pool of word candidates to the minimum set of tokens whose cumulative probability of appearing exceeds p. Through experimenting, we set the top-p value at 3 and top-k value at 0.95. We also let the model generate 20 different beams (which is a comparatively large number) resulting in more coherent and sensible text generation. Additionally, we set \textit{early\_stopping} to ‘True’ meaning that the best output is chosen only when all beams reach the ‘End Of Speech’ special token (i.e. the end of the predicted sequence) allowing a better candidate selection. \\
	
	Finally, we try to prevent a repetition of the information in the summaries, setting \textit{no\_repeat\_ngram\_size} to 3 so that no n-gram appears more than thrice in the generated sequence, and adjusting \textit{repetition\_penalty} to 2, penalizing the predicted probabilities of tokens that have already appeared in the generation. These hyperparameter settings were chosen empirically relying on the results of the evaluation metrics that are discussed below. 

\section{Evaluation metrics}

We evaluated the resulting summaries with different measures of the summarization quality. First, we compute the ROUGE score [14] that shows the n-gram overlap between the texts in terms of precision, recall, and F1 measure. We provide an F1 score for unigram and bigram overlap (ROUGE-1, ROUGE-2), and the longest common subsequence (ROUGE-L). Additionally, we calculate a BLUE score [15] – it also counts the unigram overlap between the texts, however, it penalizes words that appear in the summary more frequently than they do in the human-written text (this way, excessively repeating the words that are present in the text does not falsely increase precision). \\

	We also employ BERTscore metric [16] that computes the pairwise cosine similarity between the tokens of the original and the generated summary using contextual embeddings. BERTscore may be a more adequate measure for our task since abstractive summarization techniques can reorganize sentences, replace words with their synonyms, and so forth. BERTscore makes use of the Transformers architecture that is able to grasp the context and semantics of words, thus, reporting higher scores if the general idea of the sentences is similar even if the word match is not exact. BERTscore also provides precision, recall, and F1 estimates. 

\section{Results}

Table~\ref{metrics} displays scores achieved on the test section of Gazeta dataset. For comparison, we also included the scores of the current state-of-the-art model for summarization in Russian (mbart\_ru\_sum\_gazeta) [6] that was trained and evaluated on the same dataset that we used. 

\begin{table}[!h]
\caption{Test scores on Gazeta dataset}
	\centering
	\renewcommand{\arraystretch}{1.5}
	\begin{tabular}{ccc}
		\toprule
		    & ruGPT3 score	& mbart\_ru\_sum\_gazeta \\
		\midrule
		ROUGE-1 F1	& 11.4	& \textbf{32.1}  \\
		ROUGE-2 F1	& 1.4	& \textbf{14.2} \\
        ROUGE-L F1	& 10.0	& \textbf{27.9} \\
        BLUE	& 23.1	& \textbf{50.1} \\
        BERTscore: precision	& \textbf{0.87}	& 0.73 \\
        BERTscore: recall	& \textbf{0.90}	& 0.73 \\
        BERTscore: F1	& \textbf{0.89}	& 0.73
 \\
		\bottomrule
	\end{tabular}
	\label{metrics}
\end{table}

As can be seen, our model performs poorly when evaluating with metrics based on n-gram overlap, scoring approximately three times worse than mBART. Nonetheless, looking at BERTscore our solution shows significant improvements upon the previous results. As we have discussed above, BERTopic may be a better option for evaluating abstractive summarization, since such models can generate novel words, replace words with synonyms, rearrange the structure of the sentences, and so forth. Since the task is to detect the core ideas of the text and preserve its original message, BERTopic may provide a more reasonable measure of quality because it can gauge the meaning rather than merely score the exact matching of words. \\

	On this basis, we conclude that our model delivers promising results in text summarization.  Additionally, we validate a sample of generated summaries by hand, detecting several liabilities that can be further improved. 

\section{Analysis and Examples}

In Table~\ref{t1} we provide examples of good-quality model summarizations that come quite close to the human annotations.

\begin{table}[!h]
\caption{Examples of good model performance}
	\centering
	\selectlanguage{russian}
	\renewcommand{\arraystretch}{2}
	\begin{tabular}{p{0.45\linewidth} | p{0.45\linewidth}}
		\toprule
		    human summary & ruGPT3 summary \\
		\midrule
		Депутаты Госдумы одобрили в третьем чтении законопроект о пенсионном обеспечении граждан. Согласно документу, с 1 января 2020 года пенсии будут проиндексированы на 6,7\%. Это позволит повысить размер пенсий, повысить их размер и размер страховых взносов. & Работающие пенсионеры дождались индексации. Это следует из уже одобренных в первом чтении поправок в Конституцию. Так что работающие пенсионеры смогут получить замороженную с 2016 года прибавку к пенсии. В России их насчитывается более девяти миллионов. Эксперты называют это восстановлением справедливости.\\ 
		\hline
		20 января состоится заседание Совета Госдумы, который рассмотрит вопрос о включении в повестку нижней палаты парламента на 21 января создание должности зампредседателя Совета безопасности РФ, которую займет бывший премьер-министр Дмитрий Медведев. Некоторые эксперты трактуют это как понижение, однако на деле Медведев фактически станет правой рукой Владимира Путина в сфере безопасности и вторым лицом одного из важнейших совещательных органов России. & Дмитрий Медведев назначен на должность председателя совета безопасности при президенте РФ. Соответствующий указ был подписан президентом РФ Владимиром Путиным в ходе послания Федеральному собранию в декабре прошлого года. До этого Медведев занимал пост главы администрации президента. \\
		\hline
		Владимир Путин позвонил премьер-министру Израиля Биньямину Нетаньяху и поздравил его с юбилеем. В ходе диалога они обсудили ситуацию в Сирии и некоторые актуальные двусторонние вопросы. & Премьер-министр Израиля Биньямин Нетаньяху в ходе встречи с президентом России Владимиром Путиным заявил о готовности к диалогу с Россией по вопросу о транзите гуманитарных грузов через сирийскую территорию. Израильский премьер также выразил надежду на диалог с Москвой
	\end{tabular}
	\label{t1}
\end{table}
\newpage

\selectlanguage{english}
However, manual validation also highlights some of the issues that our model faces. First of all, it tends to replace the Named Entities given in the text (i.e. surnames, companies’ names, dates, places) with novel Named Entities. Perhaps, it is due to the fact that the model aims to paraphrase the text without taking into account that some information should not be changed. This can dramatically decrease the quality of the summarization because the model can correctly describe an event but mistakes essential details about who participated in it, or when did it happen. Examples of this behavior are shown in Table~\ref{t2}. 
\newpage

\begin{table}[!h]
\caption{Examples of Named Entities changes}
	\centering
	\selectlanguage{russian}
	\renewcommand{\arraystretch}{2}
	\begin{tabular}{p{0.45\linewidth} |     p{0.45\linewidth}}
		\toprule
		    human summary & ruGPT3 summary \\
		\midrule
		Роспотребнадзор дал новые советы россиянам в связи с распространением коронавируса COVID-19. Теперь ведомство рекомендует не пользоваться общественным транспортом в часы пик, а также воздержаться от посещения мест массовых скоплений. Кроме того, следует использовать одноразовые маски, меняя их каждые 2-3 часа, а также сократить тесные объятия и рукопожатия. & Скандально известный вирус Эбола добрался и до России. Роспотребнадзор предостерег россиян от поездок за границу в связи с эпидемией лихорадки Западного полушария, которая может распространиться по всей стране в ближайшие несколько недель.\\ 
		\hline
		В сети появились первые рецензии к сериалу «Ведьмак», премьера которого состоится на стриминговом сервисе Netflix 20 декабря. Кинокритики, получившие доступ к первым эпизодам шоу, отметили его превосходство над культовым сериалом «Игра престолов» от телеканала HBO. & Американский телеканал NBC показал первый эпизод из культовой франшизы «ведьмаков». Главную роль в нем исполнит актер Бенедикт Камбербэтч, а главные роли в других частях шоу исполнят Джейсон Стэйтем и Эмма Уотсон. \\
		\hline
		В США в продажу поступила книга Мэй Маск, матери миллиардера и основателя SpaceX Илона Маска. В книге она рассказала не только историю своего успеха, но и о том, как едва сводила концы с концами, пытаясь в одиночку прокормить троих детей. & Популярная телеведущая и бывшая модель Мэй Макларен опубликовала в своем инстаграме новую книгу, в которой рассказала о своих отношениях с миллиардером и его семье. Женщина делится своими советами и рассказывает, какие моменты стали для нее переломными. \\ 
		\hline 
		В Сети появился тизер-трейлер фильма «Чернобыль: Бездна» с Данилой Козловским в главной роли. Лента расскажет о ликвидации последствий аварии на Чернобыльской АЭС и людях, которые ценой своей жизни предотвратили глобальную катастрофу. & В российский прокат выходит первый в России сериал о Чернобыле, снятый режиссером Алексеем Козловым. Главные роли в нем исполнили актеры из Швеции, Дании, Великобритании и России. \\ 
	\end{tabular}
	\label{t2}
\end{table}

\selectlanguage{english}
Sometimes the issue is more severe, so that model confuses the actors and their actions, and misses the point of a story completely. We hypothesize that this happens in a case when there is an abundance of Named Entities that confuse the model (examples can be seen in Table~\ref{t3}). 

\newpage

\begin{table}[!h]
\caption{Examples of the model missing the key idea of the text}
	\centering
	\selectlanguage{russian}
	\renewcommand{\arraystretch}{2}
	\begin{tabular}{p{0.45\linewidth} | p{0.45\linewidth}}
		\toprule
		    human summary & ruGPT3 summary \\
		\midrule
	Австралиец Саймон Доранте-Дэй, уверенный, что он непризнанный внебрачный сын принца Чарльза и Камиллы Паркер-Боулз, решил потребовать признание этого факта через Верховный суд Австралии. Он пытается доказать, что наследник британского престола и его супруга — его настоящие родители. Накануне, 27 января, бельгийке Дельфин Боэль удалось добиться подтверждения отцовства бывшего короля Бельгии Альберта II — ей помог тест ДНК, который экс-монарх согласился пройти. & Королева Великобритании Елизавета II и принц Чарльз подали иск против своего приемного отца, британского принца Уильяма и герцогини Эдинбургской. Они обвиняют его в незаконном лишении их прав на наследование престола. \\ 
		\hline
		В сети появились первые рецензии к сериалу «Ведьмак», премьера которого состоится на стриминговом сервисе Netflix 20 декабря. Кинокритики, получившие доступ к первым эпизодам шоу, отметили его превосходство над культовым сериалом «Игра престолов» от телеканала HBO. & Американский телеканал NBC показал первый эпизод из культовой франшизы «ведьмаков». Главную роль в нем исполнит актер Бенедикт Камбербэтч, а главные роли в других частях шоу исполнят Джейсон Стэйтем и Эмма Уотсон. \\
		\hline
		Телеведущий Джеймс Корден оправдался перед зрителями за появившийся в сети ролик, на котором видно, что в рамках «Караоке в машине» шоумен вместе с поп-исполнителем Джастином Бибером едут в автомобиле, который на самом деле везет грузовик. Комик заверил, что подобное скорее является исключением из правил ради безопасности.

 & На канале CBS вышел новый выпуск популярного шоу «Карикатура», в котором ведущий Джеймсу Кэмерону задают вопросы, касающиеся его личной жизни, а также рассказывают о его отношениях с бывшей женой и друзьями. \\ 
		\hline 
		Выступая на конференции в Сан-Франциско, глава Apple Тим Кук осудил компании, занимающиеся сбором данных пользователей в качестве бизнес-модели. По словам менеджера, обмен персональной информации на доступ к технологиям — это «плохая сделка». В своей речи руководитель «яблочного» бренда также затронул другие темы, включая дружбу со Стивом Джобсом и слоган Apple «Think Different». & Apple хочет стать первой в мире компанией, продающей персональные данные пользователей iPhone и iPad. Корпорация планирует создать собственную социальную сеть, где пользователи смогут делиться своими данными с друзьями и родственниками. \\ 
	\end{tabular}
	\label{t3}
\end{table}

\selectlanguage{english}
Another issue is the repetition of the information – the model successfully detects the key thought, but instead of providing other details just repeats it several times (examples can be seen in Table~\ref{t4}). 
\newpage
\begin{table}[!h]
\caption{Examples of the repetition of information}
	\centering
	\selectlanguage{russian}
	\renewcommand{\arraystretch}{2}
	\begin{tabular}{p{0.45\linewidth} | p{0.45\linewidth}}
		\toprule
		    human summary & ruGPT3 summary \\
		\midrule
		Министр обороны США Марк Эспер заявил, что американские военные будут защищать сирийские месторождения нефти. Вашингтон намерен обороняться от «любых сил, которые будут ему угрожать», в том числе от России и сирийских военных. & Американские военные продолжают защищать нефтехранилища на юго-западе Сирии от террористов «Исламского государства». В Пентагоне заявили о продолжении работы по защите нефтяных месторождений на юге Сирии. \\ 
		\hline
		63-летний мужчина устроил стрельбу в центре Калининграда и убил супружескую пару, а также ранил их сына. Причем женщина была беременна и, как сообщают очевидцы, звала на помощь, но злоумышленник выстрелил в нее еще раз, добив ее. Сам же стрелявший пытался покончить собой и умер в реанимации. Предварительной версией конфликта стало место на местном Центральном рынке. & Стрельбу в городе Калининграде открыл неизвестный, убивший троих человек. Об этом сообщили в полиции. Очевидцы сообщают о стрельбе, которая произошла в центре города. \\
		
	\end{tabular}
	\label{t4}
\end{table}
\newpage

\selectlanguage{english}
\section{Conclusion}

In this paper, we tested ruGPT3 ability to summarize text after fine-tuning and adjusting hyperparameters. We prove that our solution outperforms the current state-of-the-art model when using a BERT-based metric, however, it scores poorly in terms of n-gram overlap between the target summary and the generated one. \\

	Our manual validation shows that the model still suffers from a number of flaws. Namely, it can change the important details of the story (surnames, dates, numbers, etc.), thus, severely decreasing the quality of summarization. Sometimes the model fails to grasp the key idea or confuses the narrative completely. Additionally, even with penalties for repetitions it still sometimes reiterates the information in the summary. \\
	
	In future work, we will aim to fix these flaws. For example, one can penalize the model specifically for altering the Named Entities labeled during training. Alternatively, one can assemble a list of all words that are not given in the source text and penalize the model for using them. In theory, this will further restrict the model, and it will deviate much less from the original document. \\
	
Other solution would be multiple sampling. Since there can be various correct summaries of the same text, we can allow model to generate several examples and then select the best option. Additionally, one can employ dynamic modeling for generating a summary that is optimal in terms of semantics and n-gam overlap [17] (i.e. via the Bellman–Ford algorithm that computes the shortest paths in a matrix). Since ruGPT3 predicts the next token, one can accelerate the generation by dropping sequences that yield insufficient quality compared to others when adding the next word, thus, iteratively filtering the ones that score the highest according to chosen metrics. It also means that if two or more sequences score equally the model would prefer the ‘less costly’ option making the resulting summaries more compressed and less repetitive. \\ 

We also plan to train a larger model (ruGPT3medium) that can process much more complex text structure and contextual details.  

\newpage
\section*{References}
\begin{enumerate}
    \item Rezazadegan, D., Berkovsky, S., Quiroz, J. C., Kocaballi, A. B., Wang, Y., Laranjo, L., Coiera, E. W: Automatic speech summarisation: A scoping review. CoRR abs/2008.11897, 1–21 (2020).
    \item Zhang, J., Zhao, Y., Saleh, M., Liu, P. J.: PEGASUS: pre-training with extracted gap-sentences for abstractive summarization. CoRR abs/1912.08777, 1–55 (2019).
    \item Gavrilov D., Kalaidin P., Malykh V.: Self-attentive Model for Headline Generation. In: Azzopardi L., Stein B., Fuhr N., Mayr P., Hauff C., Hiemstra D. (eds.) Advances in Information Retrieval. ECIR 2019. Lecture Notes in Computer Science, vol. 11438, pp. 87–93. Springer, Cham (2019). 
    \item Vaswani, A., Shazeer, N., Parmar, N., Uszkoreit, J., Jones, L., Gomez, A. N., . . . Polosukhin, I.: Attention is All You Need. In: I. Guyon et al. (eds.) Advances in neural information processing systems. vol. 30, pp. 5998-6008. Curran Associates, Inc., NY (2017). 
    \item Bukhtiyarov A., Gusev I.: Advances of Transformer-Based Models for News Headline Generation. In: Filchenkov A., Kauttonen J., Pivovarova L. (eds) Artificial Intelligence and Natural Language. AINL 2020. Communications in Computer and Information Science, vol. 1292, pp. 54–61. Springer, Cham (2020). 
    \item Gusev I.: Dataset for Automatic Summarization of Russian News. In: Filchenkov A., Kauttonen J., Pivovarova L. (eds) Artificial Intelligence and Natural Language. AINL 2020. Communications in Computer and Information Science, vol. 1292, pp. 122–134, Springer, Cham (2020). 
    \item Rush, A.M., Chopra, S., Weston, J.: A neural attention model for abstractive sentence summarization. In: Empirical Methods in Natural Language Processing. pp. 379–389 (2015). 
       \item Raffel, C., Shazeer, N., Roberts, A., Lee, K., Narang, S., Matena, M., Zhou, Y., Li, W., Liu, P. J.: Exploring the limits of transfer learning with a unified text-to-text transformer. Journal of Machine Learning Research 21, 1–67 (2020). 
    \item Lewis, M., Liu, Y., Goyal, N., Ghazvininejad, M., Mohamed, A., Levy, O., ... Zettlemoyer, L.: Bart: Denoising sequence-to-sequence pre-training for natural language generation, translation, and comprehension. CoRR  abs/1910.13461, 1–10 (2019). 
     \item Radford, A., Wu, J., Child, R., Luan, D., Amodei, D., Sutskever, I.: Language models are unsupervised multitask learners. OpenAI Blog 1(8), 1–24 (2019).  
     \item Kieuvongngam V., Tan B., Niu Y.: Automatic text summarization of covid-19 medical research articles using bert and gpt-2. CoRR abs/2006.01997, 1–13 (2020).
     \item Shatilov A. A., Rey A. I: Sentence simplification with ruGPT3. In: Computational Linguistics and Intellectual Technologies: Proceedings of the International Conference “Dialogue 2021”, pp. 1–13, Moscow (2021). 
     \item Russian GPT3 models, https://github.com/sberbank-ai/ru-gpts,  last accessed 2021/07/21.
     \item Lin, C.: ROUGE: A package for automatic evaluation of summaries. In: Text Summarization Branches Out: Proceedings of the ACL-04 Workshop, pp. 74–81. Association for Computational Linguistics, Barcelona (2004). 
     \item Papineni, K., Roukos, S., Ward, T., Zhu, W. J.: BLEU: a method for automatic evaluation of machine translation. In: Proceedings of the 40th Annual Meeting on Association for Computational Linguistics, pp. 311–318. Association for Computational Linguistics, Philadelphia (2002).
     \item Zhang, T., Kishore, V., Wu, F., Weinberger, K. Q., Artzi, Y.: Bertscore: Evaluating text generation with bert. CoRR abs/1904.09675, 1–43 (2019).  
     \item Karpov, D.A., Struchenkov, V.I.: Dynamic programming in applied tasks which are allowing to reduce the options selection. Russian Technological Journal 8(4), 96–111 (2020).
\end{enumerate}

\end{document}